\documentclass[journal,12pt,onecolumn]{IEEEtran}
\IEEEoverridecommandlockouts
\usepackage{flushend}
\usepackage{balance}
\usepackage{cite}
\usepackage{amsmath,amssymb,amsfonts}
\usepackage{algorithmic}
\usepackage{graphicx}
\usepackage{textcomp}
\usepackage{xcolor}
\usepackage{algorithm}
\usepackage{comment}
\usepackage{tabu}
\usepackage{multirow}
\usepackage{hyperref}
\usepackage[caption=false, font=footnotesize]{subfig}

\usepackage[linesnumbered,lined,boxed,commentsnumbered,algo2e]{algorithm2e}
\def\BibTeX{{\rm B\kern-.05em{\sc i\kern-.025em b}\kern-.08em
    T\kern-.1667em\lower.7ex\hbox{E}\kern-.125emX}}

\ifCLASSOPTIONcompsoc
\else
\fi

\emergencystretch=\maxdimen
\begin{document}
%

\title{A Maritime Industry Experience for Vessel Operational Anomaly Detection: Utilizing Deep Learning Augmented with Lightweight Interpretable Models}

\author{\IEEEauthorblockN{Mahshid Helali Moghadam\IEEEauthorrefmark{1}, Mateusz Rzymowski\IEEEauthorrefmark{2}, Lukasz Kulas\IEEEauthorrefmark{2}}\\
\IEEEauthorblockA{\IEEEauthorrefmark{1}
\textit{M\"{a}lardalen University}, Sweden \\
mahshid.helali@mdu.se}\\
\IEEEauthorblockA{\IEEEauthorrefmark{2}
\textit{Gdansk University of Technology}, Gdansk, Poland\\
\{mateusz.rzymowski, lukasz.kulas\}@pg.edu.pl}}


\maketitle

\begin{abstract}
This study presents an industry experience showcasing a vessel operational anomaly detection approach that utilizes semi-supervised deep learning models augmented with lightweight interpretable surrogate models, applied to an industrial sensorized vessel, called TUCANA. We leverage standard and Long Short-Term Memory (LSTM) autoencoders trained on normal operational data and tested with real anomaly-revealing data. We then provide a projection of the inference results on a lower-dimension data map generated by t-distributed stochastic neighbor embedding (t-SNE), which serves as an unsupervised baseline and shows the distribution of the identified anomalies.
We also develop lightweight surrogate models using random forest and decision tree to promote transparency and interpretability for the inference results of the deep learning models and assist the engineer with an agile assessment of the flagged anomalies. The approach is empirically evaluated using real data from TUCANA. The empirical results show higher performance of the LSTM autoencoder---as the anomaly detection module with effective capturing of temporal dependencies in the data---and demonstrate the practicality of the lightweight surrogate models in providing helpful interpretability, which leads to higher efficiency for the engineer's decision-making.
\end{abstract}


%
\IEEEpeerreviewmaketitle

\section{Introduction}
Maritime industry, as a key player in providing cost-effective transportation for people and goods \cite{IMO}, is constantly evolving by incorporating automation, sensorization, and advanced smart functionalities into vessels to improve safety and meet the changing needs \cite{negenborn2023autonomous}.
Sensorized vessels leverage advanced sensing and communication technologies to optimize vessel performance, enhance safety, and reduce environmental impact. The vessels can be equipped with various types of sensors such as GPS, radar, Lidar, sonar, and cameras to collect data of the vessel's operations. By using this data, vessels can make more informed decisions about navigation, speed, fuel consumption, and other critical aspects of their operations, which can lead to cost savings and improved efficiency 
\cite{negenborn2023autonomous, babica2020digitalization, marine_digital}.

This study presents an industry-applied operational anomaly detection framework helping the engineer with further insights into the anomalous operational states of a sensorized functional vessel, TUCANA---sailing  between Gdansk and Gdynia in Poland. The proposed approach utilizes semi-supervised deep learning (DL) models for anomaly detection along with interpretable surrogate models to promote interoperability and transparency of the inference results---also aligned with EU AI Act \cite{AIAct}. 
We leverage standard and LSTM autoencoders (AE) as the DL models. The AE models are trained based on the log data presenting mainly the normal operational states of the vessel. They reconstruct the input data based on the learned patterns representing the normality of operational behavior. During the inference, if the reconstruction error for a data instance is significantly high, it implies that the data instance deviates from the patterns learned during training---marked it as an anomaly. In this study, the 95\textsuperscript{th} percentile of the reconstruction error distribution on the training dataset is used to determine a threshold based on which the AE model can flag the data instances with high reconstruction error as anomalies. To minimize missed anomalies, data instances with reconstruction errors near the threshold (within a small vicinity of the threshold) are also regarded as potential anomalies in this approach. The AE models are tested with two separate test datasets containing the operational log data during the occurrence of a propeller failure and performing a number of critical (stress) maneuvering test scenarios on the vessel. 

The interpretable models explain the inference of the DL anomaly detection model and articulate the reasoning behind the detection. In this regard, a random forest is trained on the output of the DL model, and significant data features influencing the inference result are extracted and identified. Based on these features an optimal decision tree (DT) model is generated to represent a surrogate interpretable model for the inference results of the DL model. The depth of the DT model can be adjusted, which allows for understanding the logic underlying the anomaly detection instances. The DT also provides the possibility of automated rule generation---intended for translating the decision-making process into human-readable rules.  


To help the engineer assess anomalies flagged by the DL model, the framework is also augmented with a projection of the inference result on a lower-dimensional data map provided by t-SNE (t-Distributed Stochastic Neighbor Embedding) \cite{van2008visualizing}. t-SNE is a non-parametric dimensionality reduction technique for visualization of high-dimensional data and outlier detection. Unlike linear dimensionality reduction techniques such as principle component analysis (PCA), t-SNE can capture non-linear relationships in the data. It reveals the underlying structure and clusters within the data. In this approach, t-SNE serves as a baseline unsupervised anomaly detection method to further help the engineer assess the anomalies identified by the DL model. 


\textit{Empirical Evaluation.} We carry out an empirical evaluation of the proposed ML-driven anomaly detection approach based on the operational data collected from TUCANA. The data consists of sensory data---collected from sensors deployed on the components of the vessel---along with the navigational data, engines' data, and voltage levels of the batteries. Besides the training data, to evaluate the performance of the anomaly detection approach on real scenarios, two test data sets containing operational log data collected from two periods in which an occurrence of a propeller failure in the vessel and several critical maneuvering scenarios happen. The evaluation is intended to assess the performance of the DL anomaly detection models, i.e., in terms of precision and recall, and also the relevance and fidelity of the rules generated from the interpretable surrogate models from the engineer's perspective. Our experimental results show that the LSTM AE model shows higher precision and recall in its detection---a precision and a recall higher than 80\% and 90\% respectively. The majority of anomalies it identifies in the test data are accurately detected and It also identifies the majority of known anomalous data points present in the test datasets. 
Regarding the lightweight interpretable model, 
as per the engineer’s review, the interpretable DT model representing the anomaly detection inference of the LSTM AE model, provides the engineer with helpful insight into reasoning, facilitates the correctness assessment of the inference, and leads to time-saving investigation of flagged anomalies. 

The rest of this paper is organized as follows: Section \ref{Sec:Industrial System Architecture} provides an overview of the system architecture and data collection setup. Section \ref{Sec:Performance_Anomaly_detection} presents the details of the proposed approach including data preprocessing, DL anomaly detection models, the interpretable surrogate models, rule generation, and the use of t-SNE. Section~\ref{Sec: Empirical_Evaluation} elaborates on the empirical evaluation including the experimental setup and the research questions that are addressed. Section \ref{Sec:Discussion} discusses the results and elaborates on the applicability of the approach from an industrial perspective. Section \ref{Sec:Related Work} provides an overview of the related work, and lastly, Section \ref{Sec:Conclusion} concludes the paper with our findings and the potential research directions for future work. 

\section{System Architecture}
\label{Sec:Industrial System Architecture}
\begin{figure*}[h]
  \centering
  \includegraphics[width=0.55\textwidth, height= 9cm]{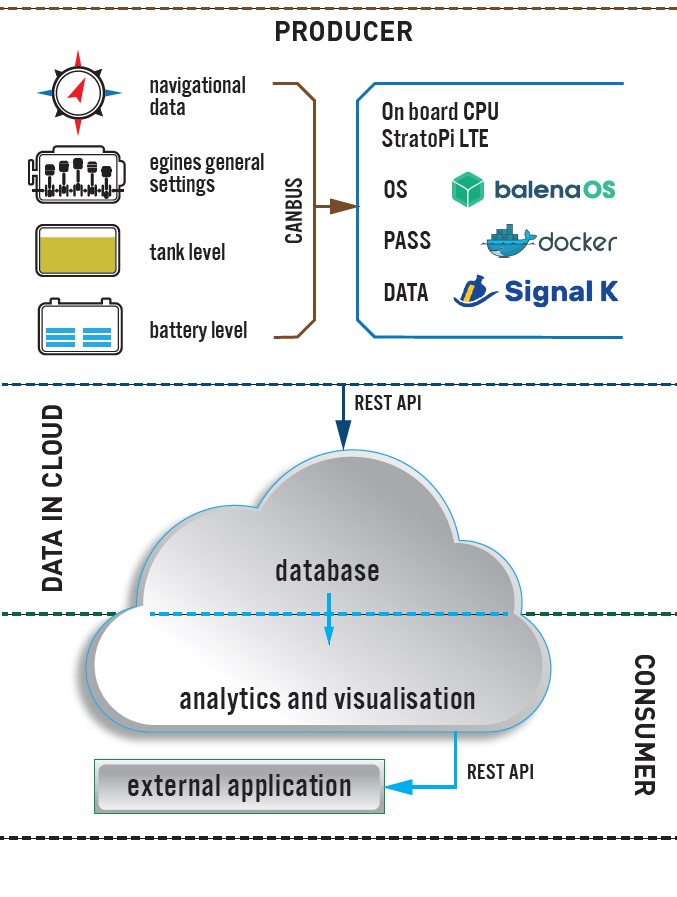}
  \caption{A scheme of data streaming from TUCANA to data storage and visualization platform.}
  \label{fig: data_streaming_strcuture}
\end{figure*}

The architecture of the data collection and vessel monitoring system is illustrated in Figure \ref{fig: data_streaming_strcuture}. Data is gathered from onboard sensors installed on various vessel components. The collection and aggregation process is facilitated by a controller area network (CAN) bus adhering to the NMEA2000 standard, managed by the onboard processing unit (OBU). The sensor network generates navigational data, including the vessel's position, direction of movement, and engine metrics such as RPM, fuel level, and battery charge. This setup for the collection of vessel's data is common in marine applications \cite{4401548,Kessler2021} and is compatible with most of the products offered by maritime equipment manufacturers.
The OBU performs data collection and aggregation using a single-board Linux-based computer called StratoPI, designed for industrial conditions. The computer is equipped with the Balena operating system, dedicated to reliable solutions utilizing Docker containers. The data is exchanged using the open-source Signal K standard, which enables communication and information sharing among devices and applications in the maritime domain. The Signal K server has been deployed as a docker container, allowing for convenient and scalable deployment. It utilizes a representational state transfer (REST) API, enabling clients to interact with the Signal K server using standard HTTP methods.

\begin{figure*}[h]
  \centering
  \includegraphics[width=.9\textwidth, height= 8.5cm]{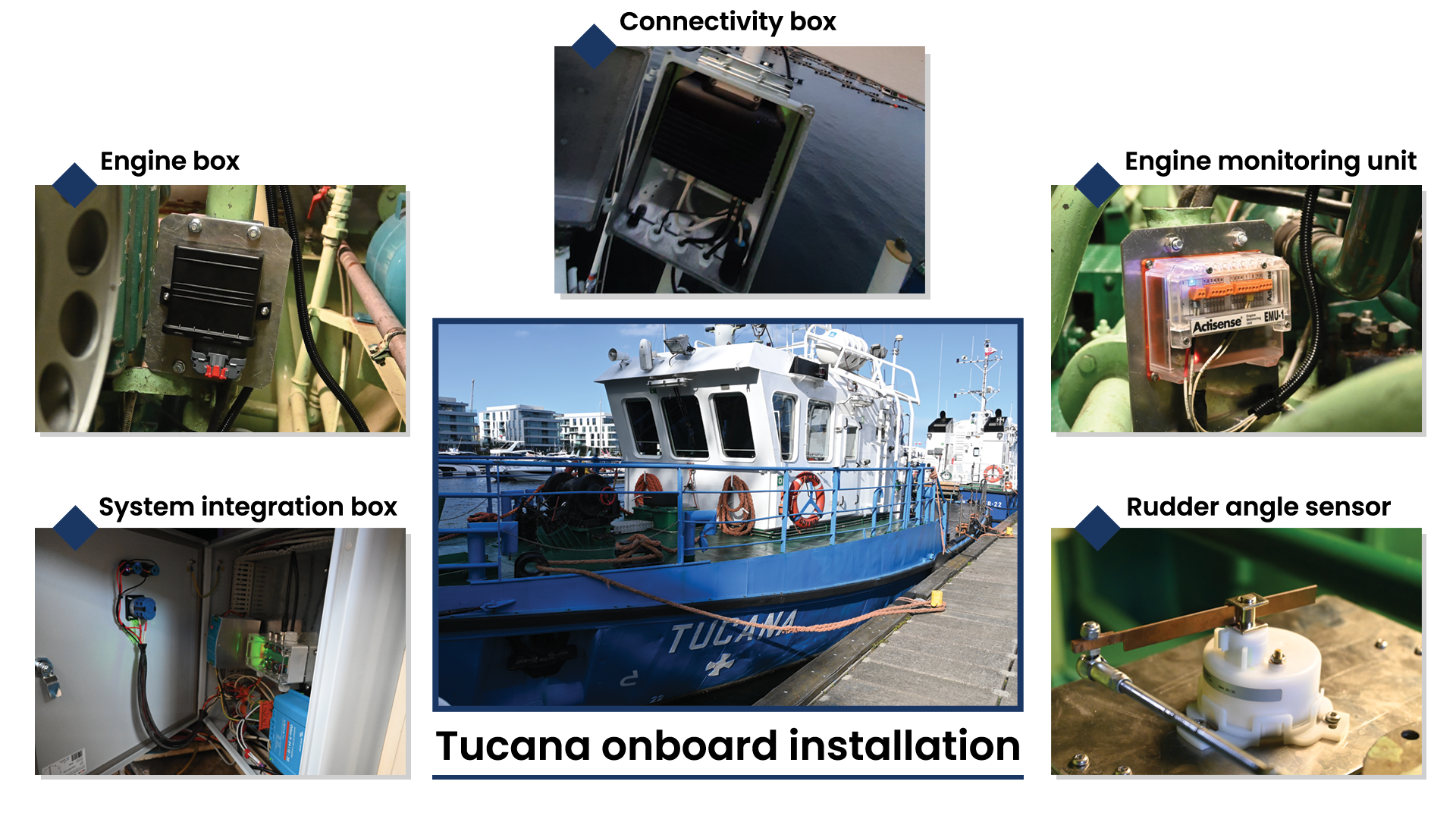}
  \caption{Overview of the TUCANA onboard installation}
  \label{fig:TUCANA_Physical_Components}
\end{figure*}

The aggregated data is wirelessly transmitted to the cloud using an LTE Cat 6 modem. As the vessel operates within an area covered by LTE base stations, there is no need for an additional dedicated wireless link. Additionally, the aggregation mechanism is designed in such a way that data synchronization with the cloud application occurs when the connection is available, which is acceptable for this application due to long-term data analysis and prediction.
Cloud-based storage is used to handle large volumes of vessel data. To ensure scalability and reliability, an InfluxDB database stores the vessel data and Grafana is used for data integration and visualization of both raw and processed data. Several templates have been proposed and can be used by domain engineers to monitor the vessel condition. Figure \ref{fig:TUCANA_Physical_Components} presents an overview of the TUCANA onboard installation and the physical components involved in the setup.  

\section{ML-driven Operational Anomaly Detection Augmented with Interpretability}\label{Sec:Performance_Anomaly_detection}
In this industry experience, the overarching objective is to develop an ML-driven operational anomaly detection approach, which identifies anomalous operational states. It is an analytics decision support system that helps the maritime engineer gain insight into the functional state of the system. The intended requirements for the anomaly detection component are mainly focused on its effectiveness in detecting operational anomalies accurately---it means fewer false alarms and missed anomalies result in better performance. The following sections elaborate on the details of the approach in different parts including data management, DL model development, and the interpretable surrogate models, which promote interpretability and also enable edge-based deployment of the anomaly detection component.

\subsection{{Data Management}}
\begin{figure*}[h]
\centering
\includegraphics[width=.7\textwidth, height=6cm]{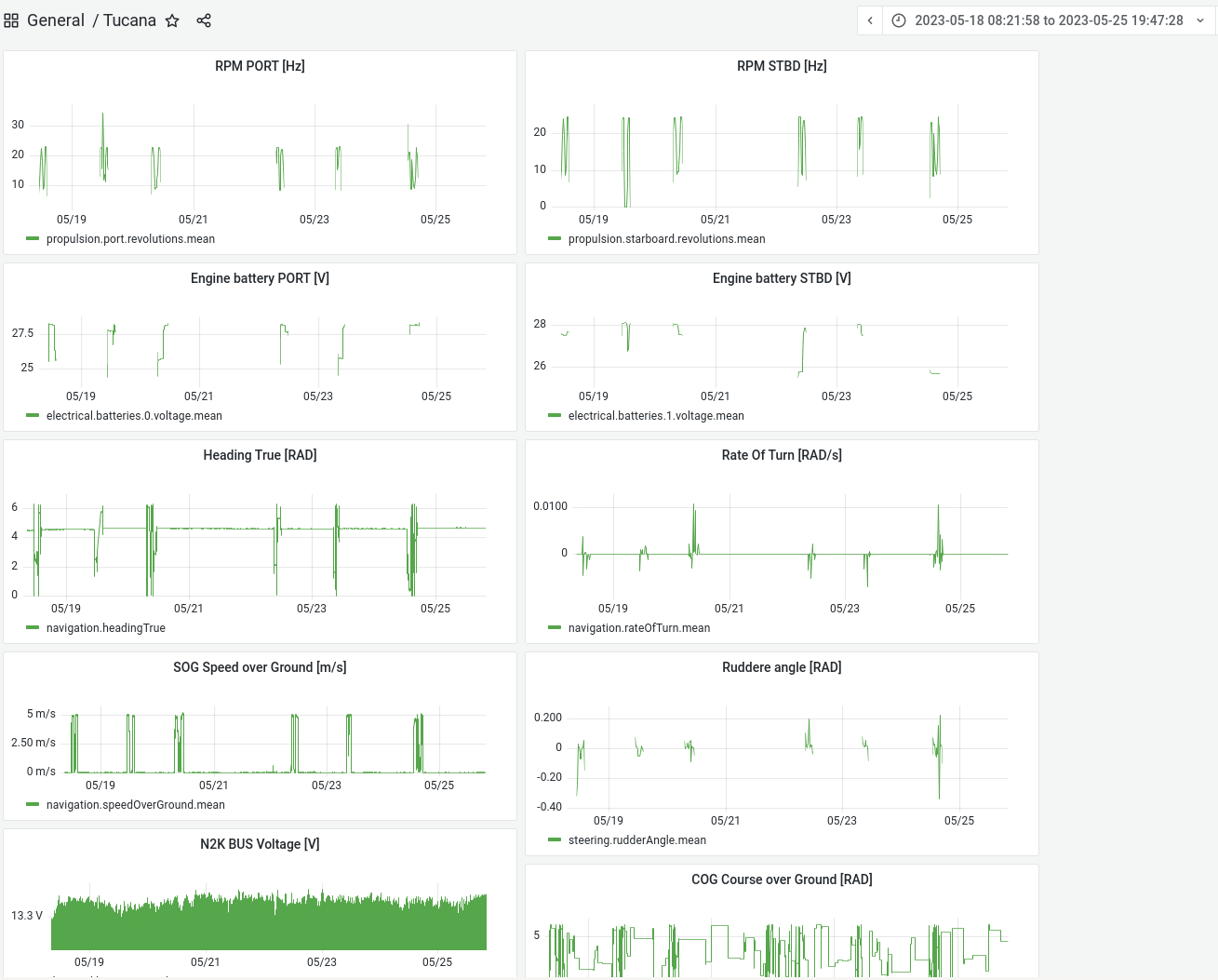}
\includegraphics[width=.7\textwidth, height=4cm]{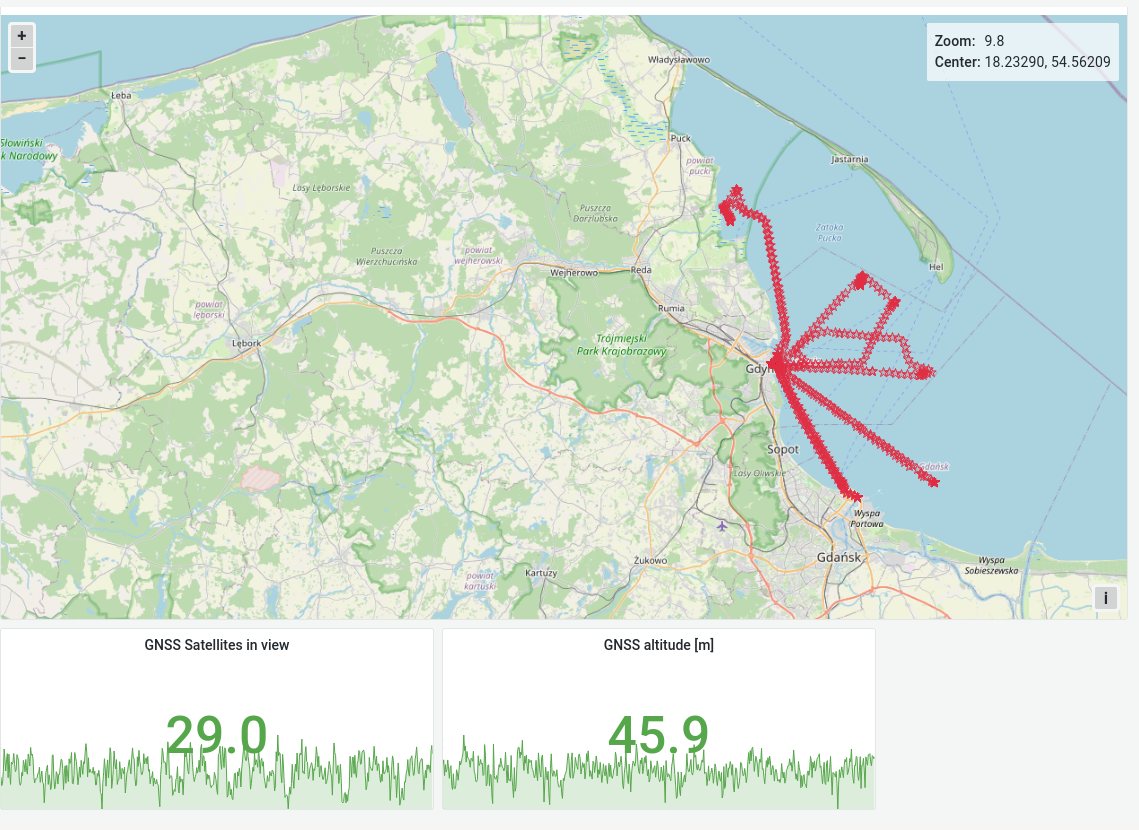}
\caption{A daily view of the data signals}
\label{fig: Data_View}
\end{figure*}

The vessel log data used for the model development is composed of the collected values for 8 signals measured by TUCANA onboard sensors. It is fetched from the Grafana dashboard---Figure \ref{fig: Data_View} presents a daily view of the recorded measurements for these signals. Each signal represents a feature in the data set. The values of the signals are time series data, though recorded at different frequencies. The signals are the revolutions per minute (RPM) of the engines (both at the port and starboard side, STBD), the voltage levels of the engines' batteries, heading true, rate of turn, speed over the ground (SOG), and rudder angle. Heading true refers to the direction in which a vessel is pointing relative to true north. It is measured in radians, with 0 indicating a heading directly towards true north. The rate of turn is determined by observing how quickly a vessel changes its heading and is expressed in radians per minute. It is an essential parameter for controlling the vessel maneuverability and maintaining the desired course or heading. Speed over the ground for a vessel refers to the actual speed at which the vessel is moving relative to the Earth's surface. It is measured in meters per second (m/s). SOG represents the vessel's speed and direction in relation to fixed points on land, while the rudder angle refers to the position of the vessel rudder relative to its neutral or centered position. It is measured in radians and indicates how much the rudder has been deflected from its neutral position. Positive values indicate deflection to the right or starboard side, while negative ones indicate deflection to the left or port side. The value recording for data signals takes place during the operation time of the vessel. 

\textbf{Pre-processing.} The signals are recorded at different frequencies, so the first step in pre-processing involves resampling the signal values to ensure a consistent frequency across all signals. In this regard, for each feature, the mean values of the signal in 10-second intervals are considered---according to a sensible granularity for the measurements. Averaging over intervals can reduce noise in high-frequency signals and is computationally efficient. Then, the empty records with null values for all the features, which represent the non-functioning period of the vessel, are removed. Additionally, the dataset is purged of the records representing the non-moving state of the vessel, i.e., floating on the surface of water---with null values for the mean voltage of the engine batteries at both port and STBD and also very small values for SoG.  Applying normalization to rescale the data values to the range $(0, 1)$ is the next step  before feeding the data to the deep learning anomaly detection models.   

\subsection{Deep Learning Models} 
We employ both vanilla (standard) AE and LSTM AE as DL models widely used for anomaly detection purposes. AE \cite{hinton2006reducing} is a neural network architecture that consists of two parts: an encoder and a decoder. The encoder takes the input data and transforms it into a lower-dimensional representation, while the decoder takes the encoded representation and reconstructs the original input data. 

In LSTM AE, the LSTM layers \cite{gers2000learning}, used in the encoder and decoder, process data in a sequence-based format and can capture the temporal dependencies in the data by using a memory cell that can selectively forget or remember information over time. In an LSTM AE, the input data fed into the encoder is one sequence, and the LSTM layer in the encoder processes the sequence while retaining information from previous elements. The output of the encoder is then fed into the decoder, which also uses LSTM layers to decode the sequence back into its original form.

AEs are used for anomaly detection based on their ability to reconstruct input data. The idea involves training the AE on a dataset containing normal (anomaly-free) data. It learns to reconstruct the normal input data by encoding it into a lower-dimensional representation and then decoding it back to the original dimensions. During the training process, the objective is to minimize the reconstruction error, which is the difference between the original input and the reconstructed output. 
During the inference, if the reconstruction error for a data instance is significantly high, it suggests that the data instance is anomalous or deviates from the patterns learned during training. The assumption is that the AE will struggle to reconstruct anomalous data, as it has not been trained on such patterns. By setting a threshold on the reconstruction error, instances with higher reconstruction errors than the threshold can be classified as anomalies. 
The choice of threshold is important, as it affects the labeling of anomalies. In this study, w.r.t the scarcity of anomalies, the threshold is determined using a common statistical approach. Specifically, the 95\textsuperscript{th} percentile of the reconstruction error distribution on the training set is used to ensure that only the most extreme deviations from normal patterns are classified as anomalies. This approach is often leveraged as a suitable baseline for anomaly detection. 
Other empirical techniques like using a separate labeled test set, experimenting with various threshold values, and evaluating the performance in terms of relevant metrics (e.g., precision, recall, and F1-score) can be used as well. Here, we benefit from using the 95\textsuperscript{th} percentile of the reconstruction error distribution to identify the cutoff point. Moreover, to minimize missed anomalies and avoid overly rigid classifications, instances with reconstruction errors near the threshold (within a small vicinity of the threshold) are regarded as potential anomalies--It is worth including them in the engineer's investigation.

\textbf{t-SNE.} To pave the way for the engineer to assess the identified anomalies and as a baseline for anomaly detection, the output of the DL model is projected on a lower-dimensional presentation of data generated using t-SNE. It computes the pairwise similarity of the data points in the high-dimensional space using a Gaussian probability distribution. Then, it initializes points in the low-dimensional space, computes the pairwise similarity in the low-dimensional space using a Student’s t-distribution, and minimizes the KL Divergence between high- and low-dimensional distributions. It updates low-dimensional points using gradient descent. t-SNE shows the clusters within the data and reflects the distribution and relative position of the anomalies identified by the DL model.

\subsection{Interpretable Surrogate Model} 
Interpretability in machine learning \cite{molnar2020interpretable} refers to the ability to understand how a model makes predictions or decisions. The goal of interpretability is to provide insights into how a model works, identify potential biases, and build trust in the model outputs. Deep learning models are generally considered less interpretable than traditional machine learning models such as logistic and linear regression, support vector machine, and decision trees. Deep learning models typically consist of multiple layers with many parameters, and the interactions between these layers can be difficult to understand. 
In the proposed approach to satisfy the requirements of edge deployment (i.e., resource-constrained environment) for the anomaly detection model as well as provide insights into how the DL model makes decisions, we create interpretable surrogate models based on the outcome of the anomaly detection DL model---which approximates the output of the DL model. 

Decision tree is a popular and interpretable machine learning model \cite{molnar2020interpretable} that builds tree-like structures where each internal node represents a decision based on a feature, and each leaf node represents a class label. 
To create a pruned non-overfitted DT model, we extract the feature importance and build the interpretable model based on the significant main features driving the decision made by the DL model. In this regard, first, we utilize a random forest model \cite{breiman2001random} that is trained on the outcome of the anomaly detection DL model. It provides feature importance scores. Random Forest is an ensemble model that combines multiple decision trees and aggregates their predictions. It is less interpretable compared to individual decision trees, though offers more reliable estimates of feature importance by aggregating the feature importance values from its constituent trees \cite{breiman2001random}.
After identifying the significant features, a DT model is trained to give an interpretable presentation of the anomaly detection performed by the DL model. 

\textbf{Rules.} The generated DT also offers a set of human-readable rules that describe how input features lead to specific outcomes.
The "if-then" structure of the rules makes it even more straightforward to grasp how the features influence the prediction and provides more transparency into the decision-making process of the model. 
Examining the rules can help identify if the model decisions align with human thoughts or known ground truths. 

\begin{figure*}[h]
  \centering
  \includegraphics[width=.85\textwidth, height= 7cm]{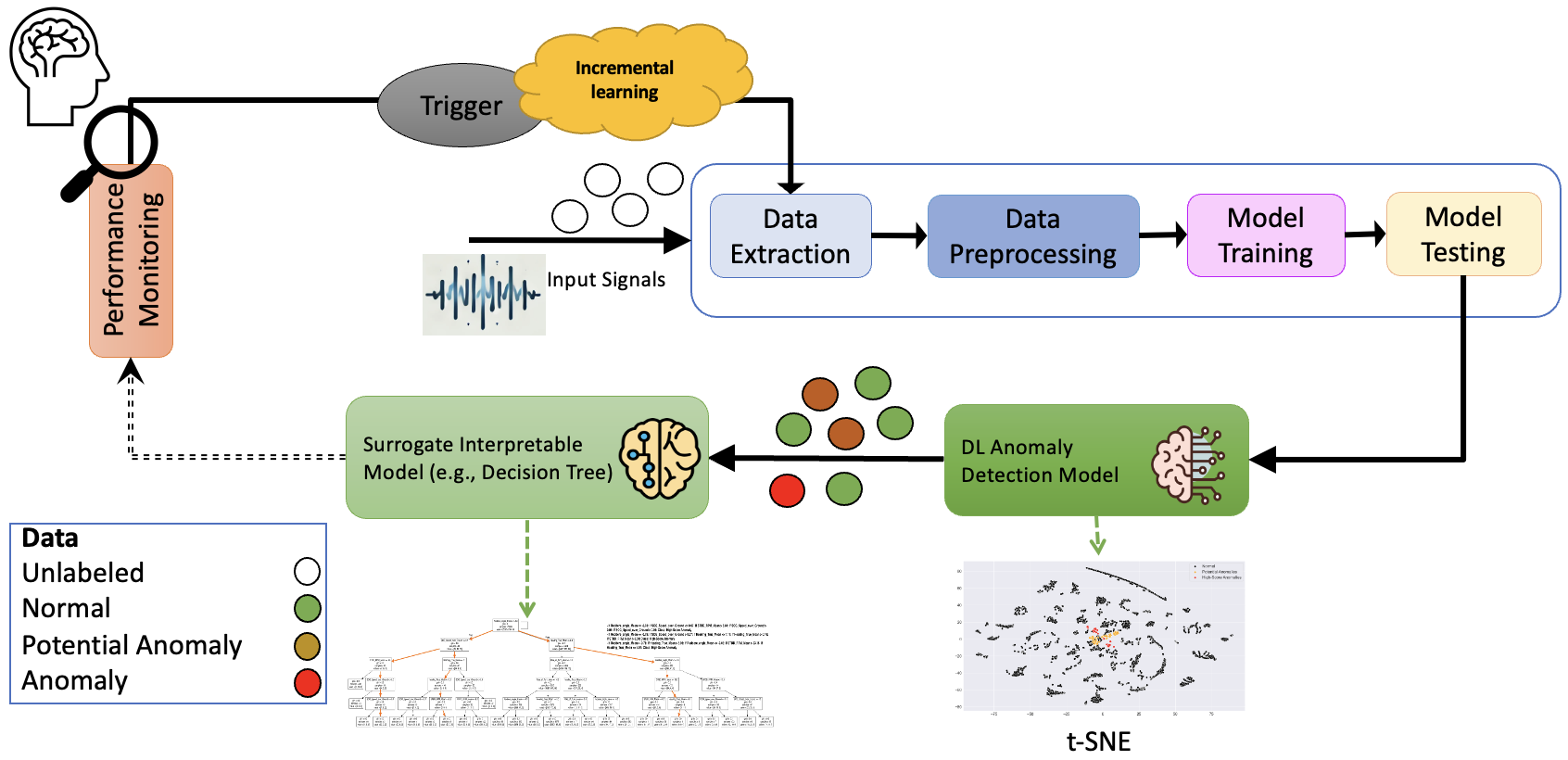}
  \caption{ML-driven operational anomaly detection augmented with lightweight interpretable models, for the TUCANA sensorized vessel}
  \label{fig:ML_Operational_anomnaly_Detection_process}
\end{figure*}

Figure \ref{fig:ML_Operational_anomnaly_Detection_process} illustrates the proposed framework of the DL-based operational anomaly detection augmented with surrogate interpretable models. The deployment setup of this framework involves a human-in-the-loop performance monitoring to track and detect performance model degradation (drift) \cite{bayram2022concept}, in which it can initiate retraining the DL model or fine-tuning it with the new data \cite{ModelRetraining}. Performance degradation of the DL model can happen, for instance, due to covariate shift in the data (changes in the distribution of input features over time) can be detected based on the increased false positives or false negatives rates in the anomaly detection---reviewed by the human in the loop.   

\section{Industrial Empirical Evaluation}\label{Sec: Empirical_Evaluation}
We perform an empirical evaluation of the proposed ML-driven operational anomaly detection approach. The experimental setup, the data used for the model development, test data, and research questions are discussed in the following subsections.

\subsection{Experimental Setup}
\textit{Data:} We fetch three months of operational data for the development of DL anomaly detection models ($90\%$ used for training and $10\%$ for validation). We use two test datasets collected from two time periods, i.e., 8-day and 11-day intervals. The first test dataset, \textit{test data 1}, is log data collected from a period in which an occurrence of propeller failure has been reported. The second test dataset, \textit{test data 2}, includes the log data collected in a period in which a number of critical maneuvering test scenarios were done on the vessel.


\textit{DL Models Architecture.} The vanilla AE is configured as follows: the encoder consists of three {Dense} layers with 32, 16, and 8 units and ReLU activation functions. 
The decoder includes two Dense layers with 16, 32 units and ReLU activation functions, and a Dense layer---as the last layer---with 8 units 
and sigmoid activation function. The output of the decoder is the reconstructed data, which should ideally match the input data.

The LSTM AE model is intended to take sequence input, so before feeding the pre-processed training data into the model, we use a sliding window technique to adjust the training dataset and construct the input sequences for the model. In the LSTM AE model, the encoder part consists of two LSTM layers with 64 and 32 units, followed by a RepeatVector layer which takes the output of the encoder LSTM layer and expands the compressed representation to match the original input sequence length. The decoder part contains two LSTM layers with 32 and 64 units as well as a TimeDistributed dense layer, which applies a dense layer to each time step of the sequence, which enables the model to output the correct dimensionality for every time step in the sequence and reconstruct the entire sequence. We consider an input sequence length of 13, with a stride set to 1. The input sequences represent intervals of $120 s$. 

For training the models, we use Adam (Adaptive Moment Estimation) Optimizer, as a commonly used optimization algorithm for training the deep learning models, and the ReLU (Rectified Linear Unit) activation function as a computationally efficient function 
We use mean squared error (MSE) to compute the reconstruction error, which measures the difference between the input and the reconstructed output and serves as the loss function during training to adjust the model parameters. By minimizing the reconstruction error, the AE learns to accurately reconstruct normal input data.

In both AE models, the batch size and the number of epochs are set to 64 and 50 respectively. We use the 95\textsuperscript{th} percentile of the reconstruction error values as the threshold for detecting anomalies. After training the AE model, the training dataset is passed through the trained model, and the MSE between the input and the reconstructed output for each data sample is calculated. After sorting the calculated MSE values in ascending order, the MSE value at the 95\textsuperscript{th} percentile is used as the threshold and the samples with the reconstruction error above the threshold are considered (potential) anomalies. The threshold for the LSTM and vanilla AEs, according to the 95\textsuperscript{th} percentile technique, is set to $0.036$ and $0.013$ respectively. Therefore, in order to provide a unified view over the identified anomalies (and potential anomalies) by both DL models, empirically we regard the points with the reconstruction error higher than $0.05$ as \textit{high-score anomaly} and the ones between the predefined thresholds and $0.05$ as \textit{potential anomaly}.

\subsection{Research Questions} 
In this industry experience paper we intend to address the following two research questions:\\
\textbf{RQ1}: How can we effectively utilize semi-supervised DL models for operational anomaly detection 
for an industrial functional vessel?\\
\textbf{RQ2}: How can we enhance transparency and interoperability to the inference of the anomaly detection DL model and also help the engineer with efficient decision-making?

\section{Results and Discussion} \label{Sec:Discussion} 
\subsection{RQ1: DL Operational Anomaly Detection} 
\begin{figure*}[h]
\centering
\subfloat[LSTM AE]{
    \label{fig:Test1_LSTM_Autoencoder_Model}
    \framebox{\includegraphics[width=0.48\linewidth, height=5cm]{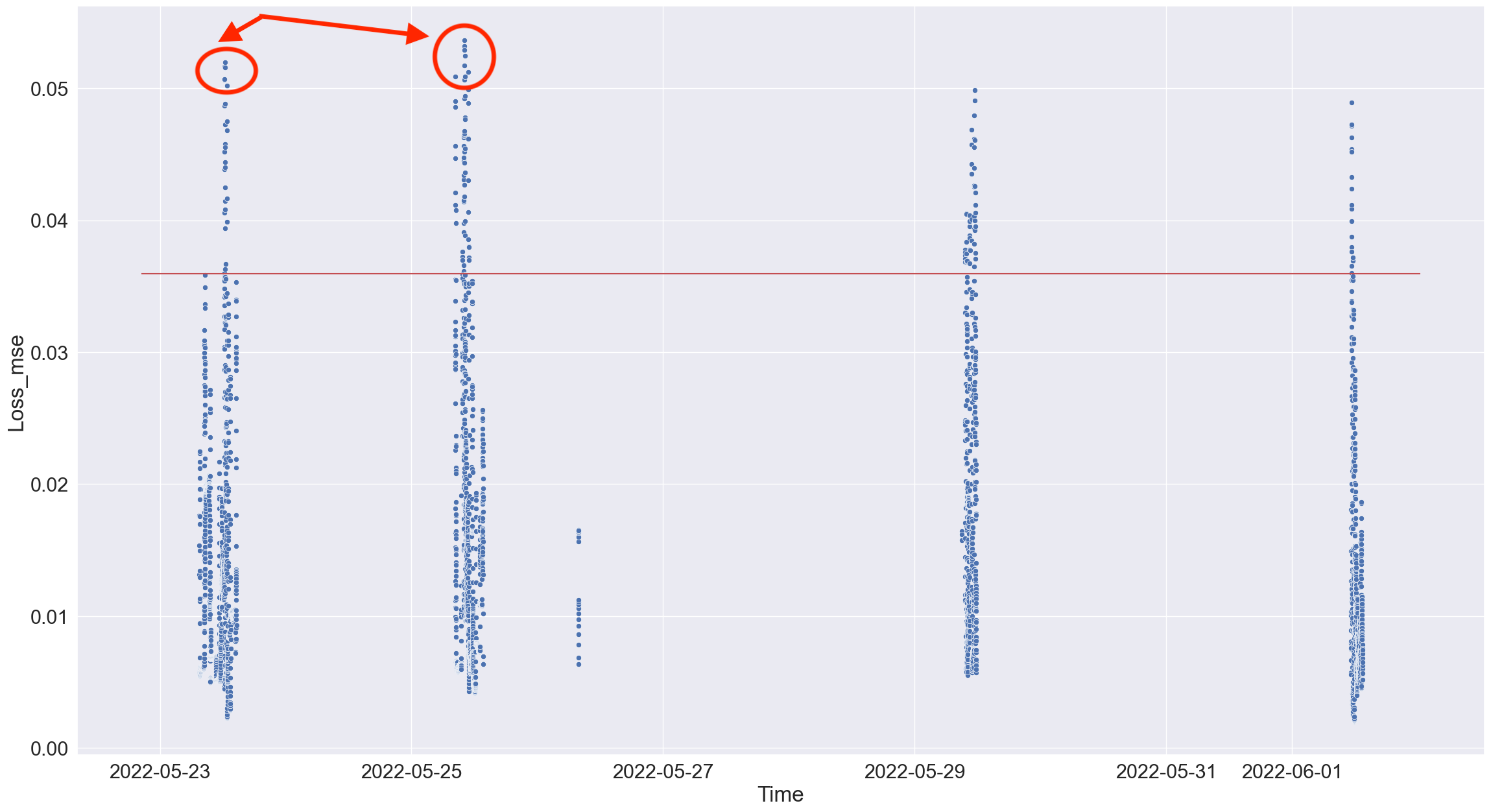}} } 
\subfloat[Vanilla AE]{
    \label{fig:Test1_Vanilla_Autoencoder_Model}
    \framebox{\includegraphics[width=0.48\linewidth, height=5cm]{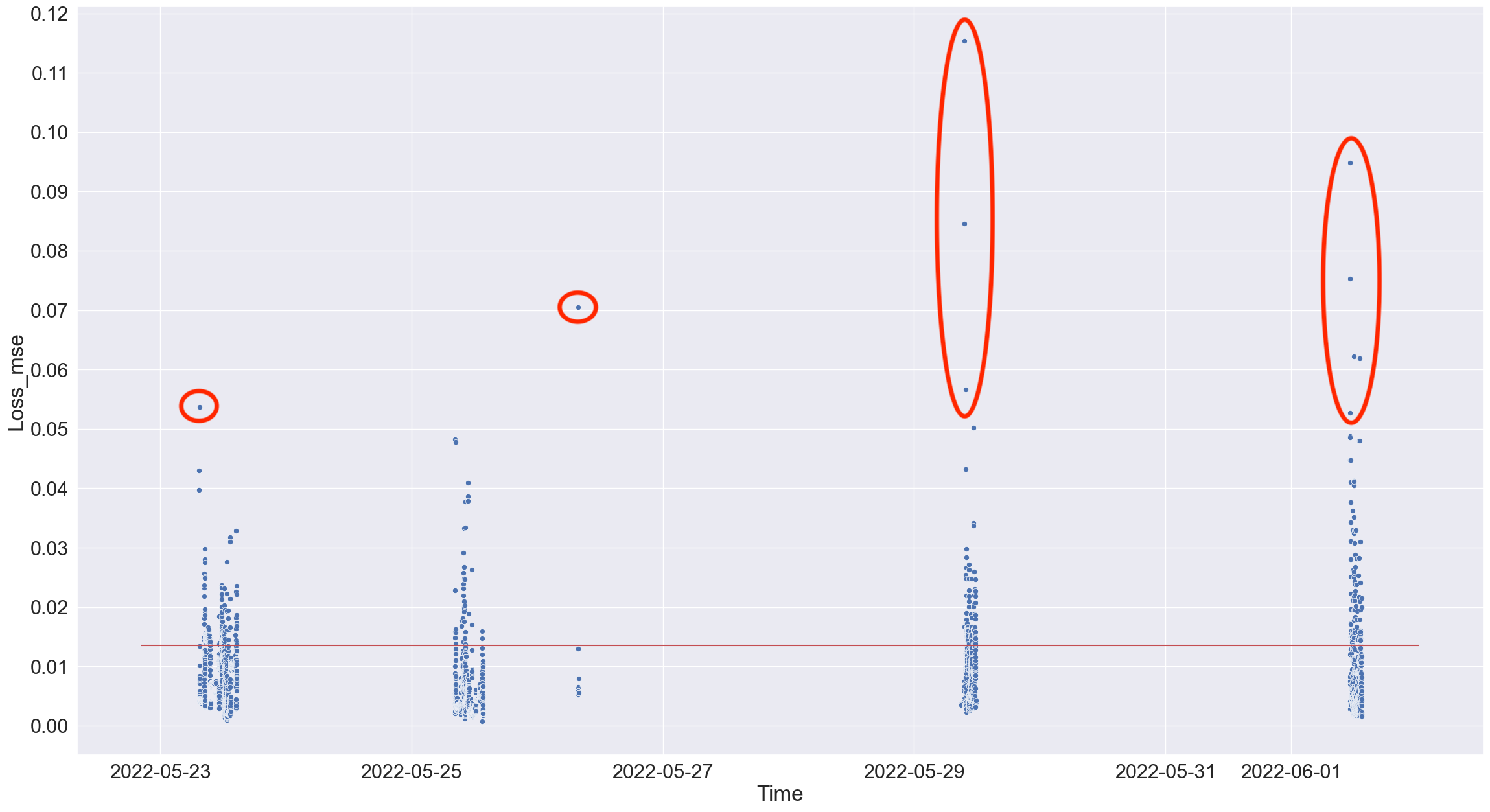}}}
\caption{Detected anomalies by the DL anomaly detection models on test data 1. }
\label{fig:Test1_DL_anomaly_detection_models}
\end{figure*}

\begin{figure*}[h]
\centering
\subfloat[LSTM AE]{
    \label{fig:Test1_LSTM_tSNE}
    \framebox{\includegraphics[width=.48\textwidth, height= 5cm]{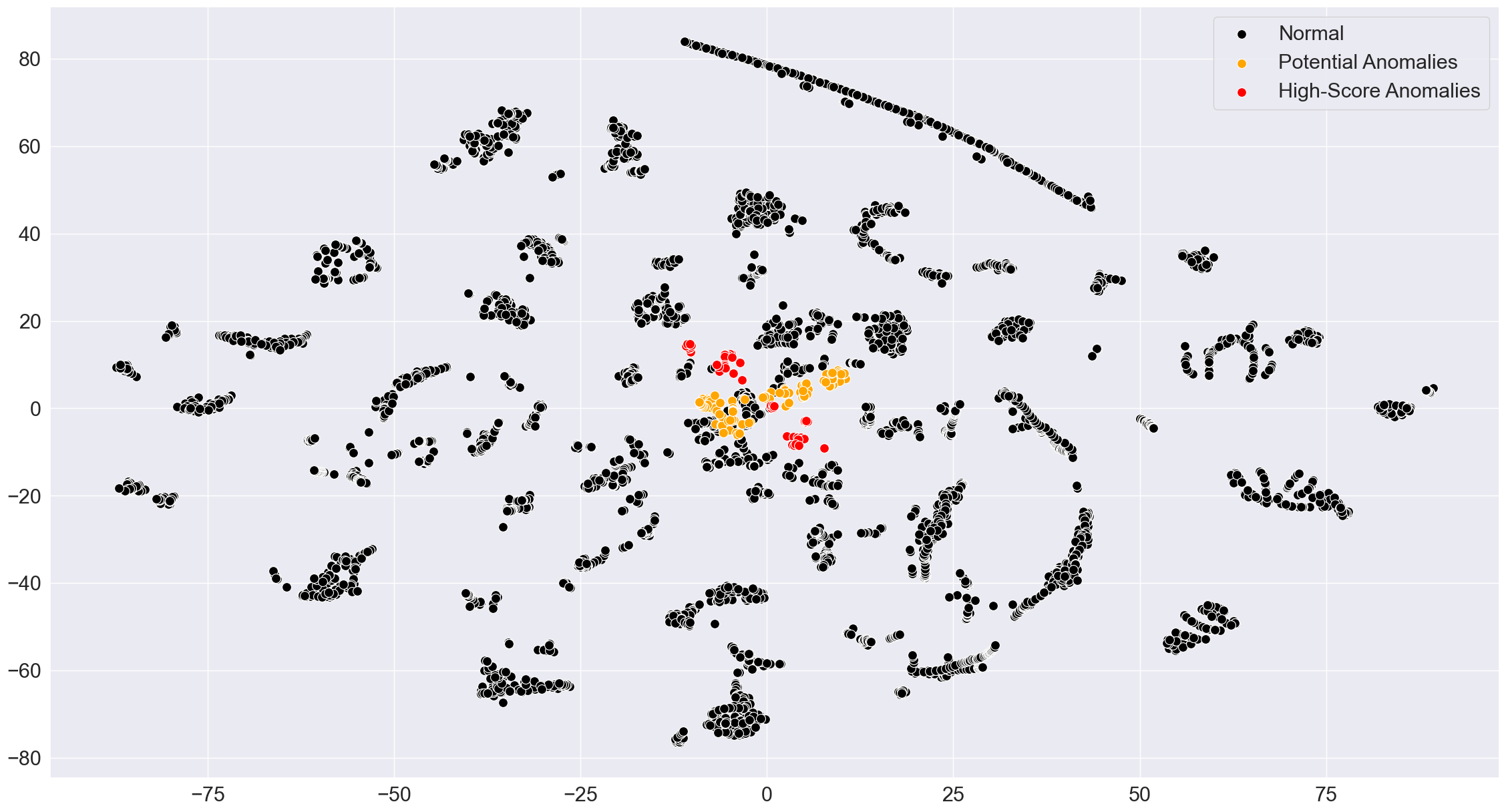}}}
\subfloat[Vanilla AE]{
    \label{fig:Test1_Vanilla_tSNE}
    \framebox{\includegraphics[width=0.48\linewidth, height=5cm]{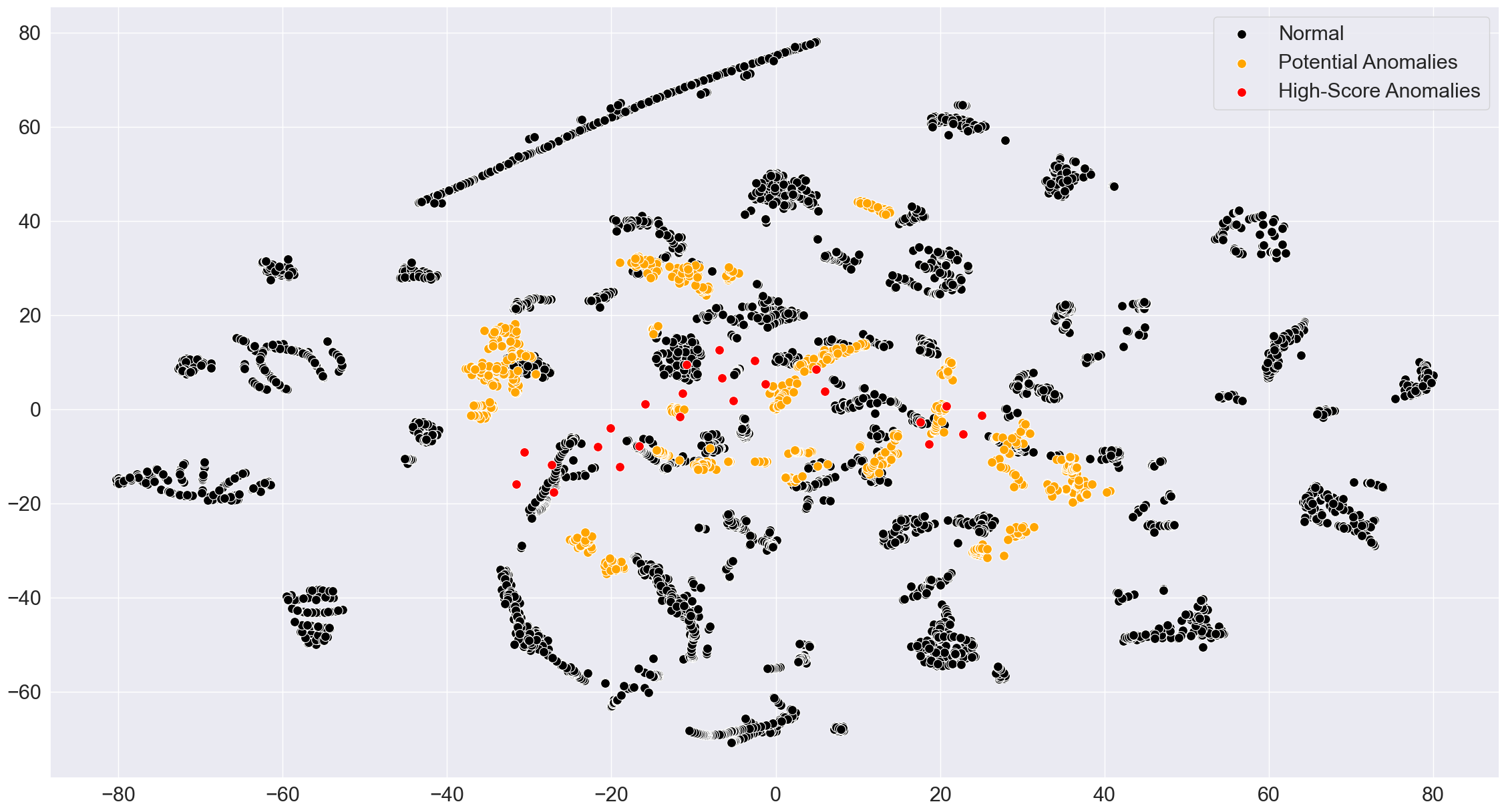}}}
\caption{t-SNE map of the test data 1 illustrating the inference result of the DL models.}
\label{fig:Test1_tSNE}
\end{figure*}

Figures \ref{fig:Test1_DL_anomaly_detection_models} shows times steps in which the DL anomaly detection models detect the occurrence of operational anomalies in test data 1. Then, to pave the way for the engineer to read and assess the result of the DL models, a t-SNE map of the data reflecting the inference output of the DL model is created. The t-SNE map provides a view of data clusters, along with the detected anomalies, and their distribution in the dataset---it is used as an unsupervised baseline for anomaly detection (Figure \ref{fig:Test1_tSNE}).

\begin{figure*}[h]
\centering
\subfloat[LSTM AE]{
    \label{Test2_LSTM_Autoencoder_Model}
    \framebox{\includegraphics[width=0.48\linewidth, height=5cm]{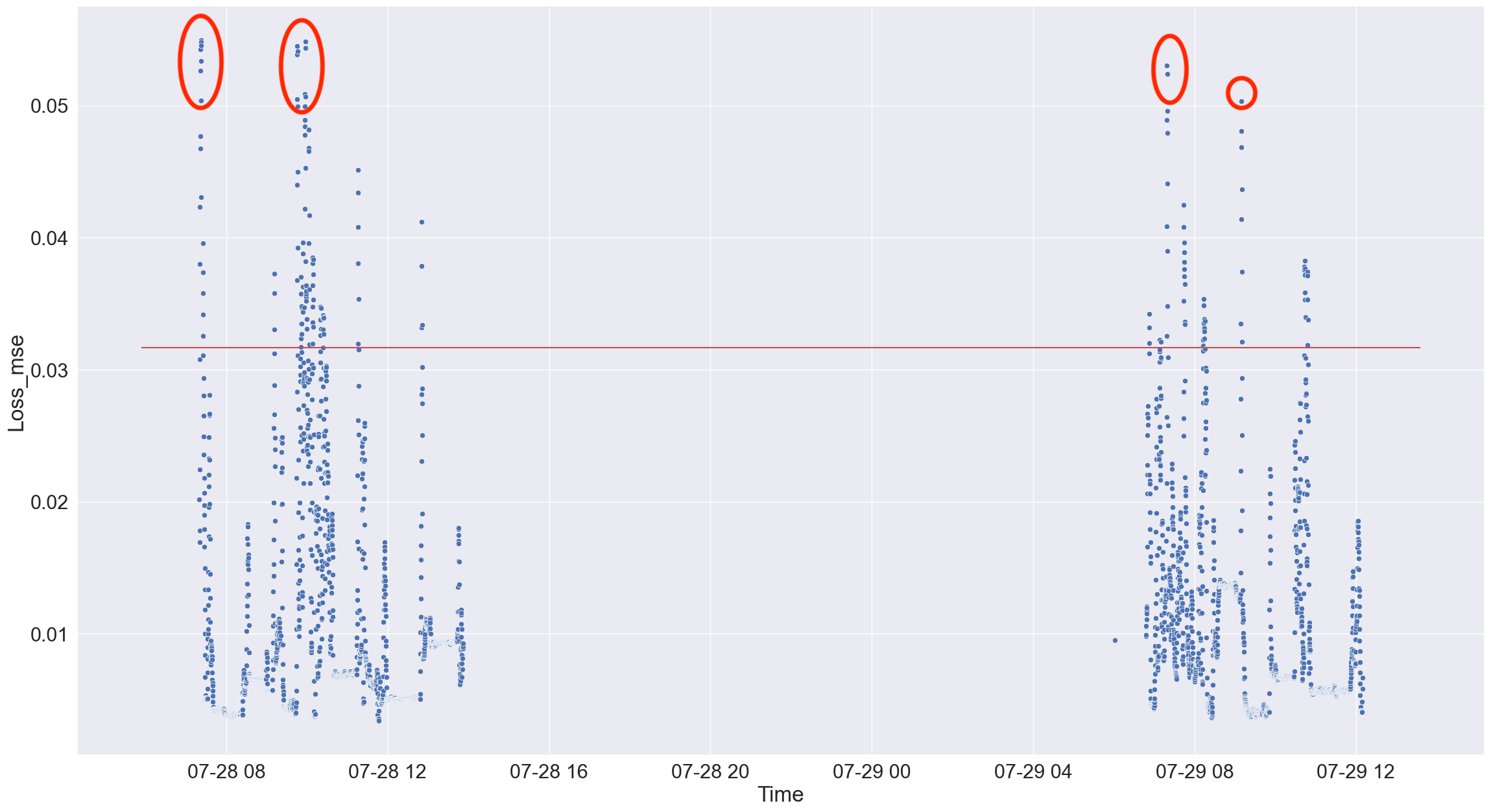}}} 
\subfloat[Vanilla AE]{
    \label{Test2_Vanilla_Autoencoder_Model}
    \framebox{\includegraphics[width=0.48\linewidth, height=5cm]{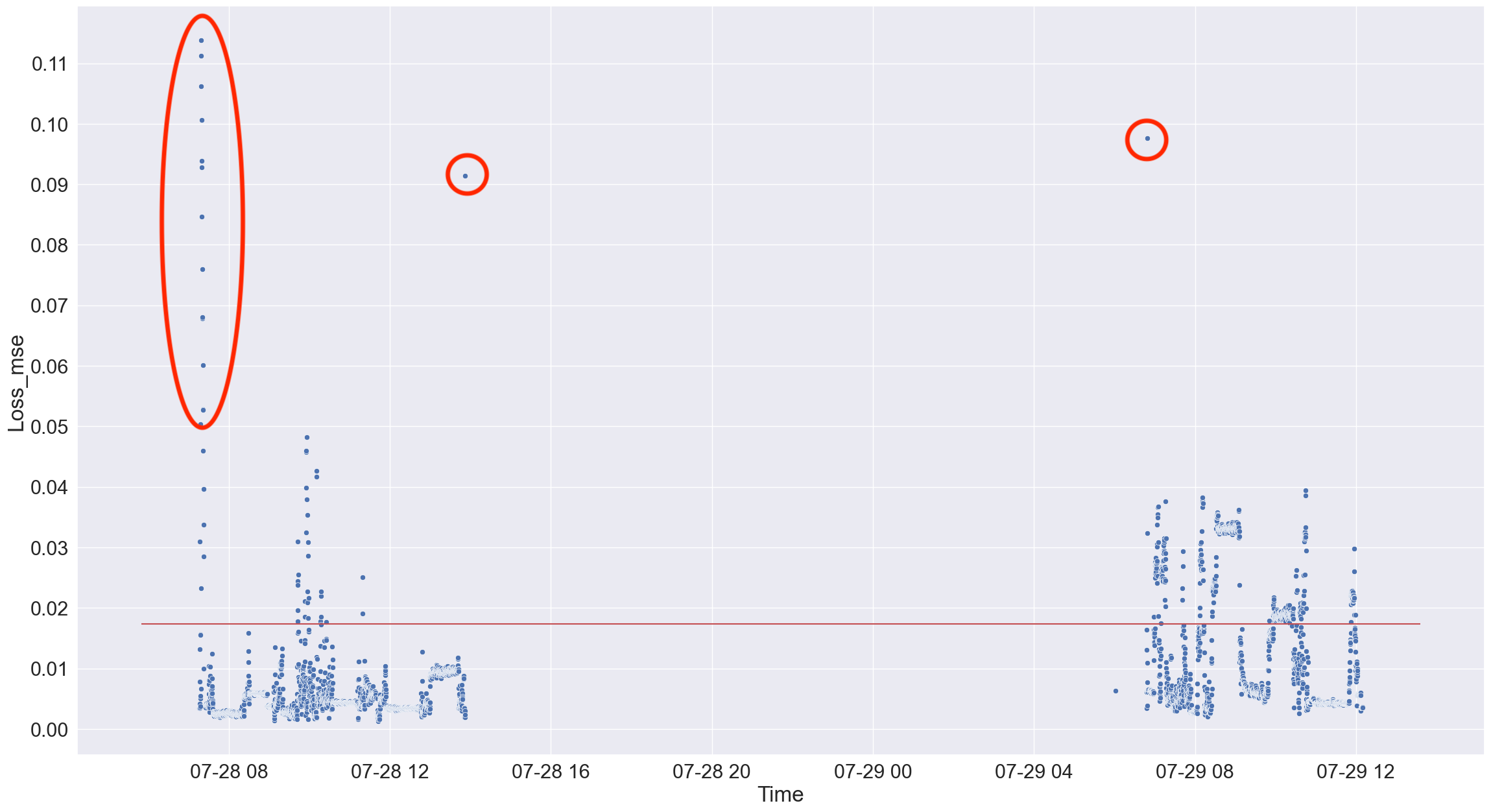}}}
\caption{Operational anomalies detected by the DL anomaly detection models on test data 2. }
\label{fig:Test2_DL_anomaly_detection_models}
\end{figure*}

\begin{figure*}[h]
\centering
\subfloat[LSTM AE]{
    \label{fig:Test2_LSTM_tSNE}
    \framebox{\includegraphics[width=.48\textwidth, height= 5cm]{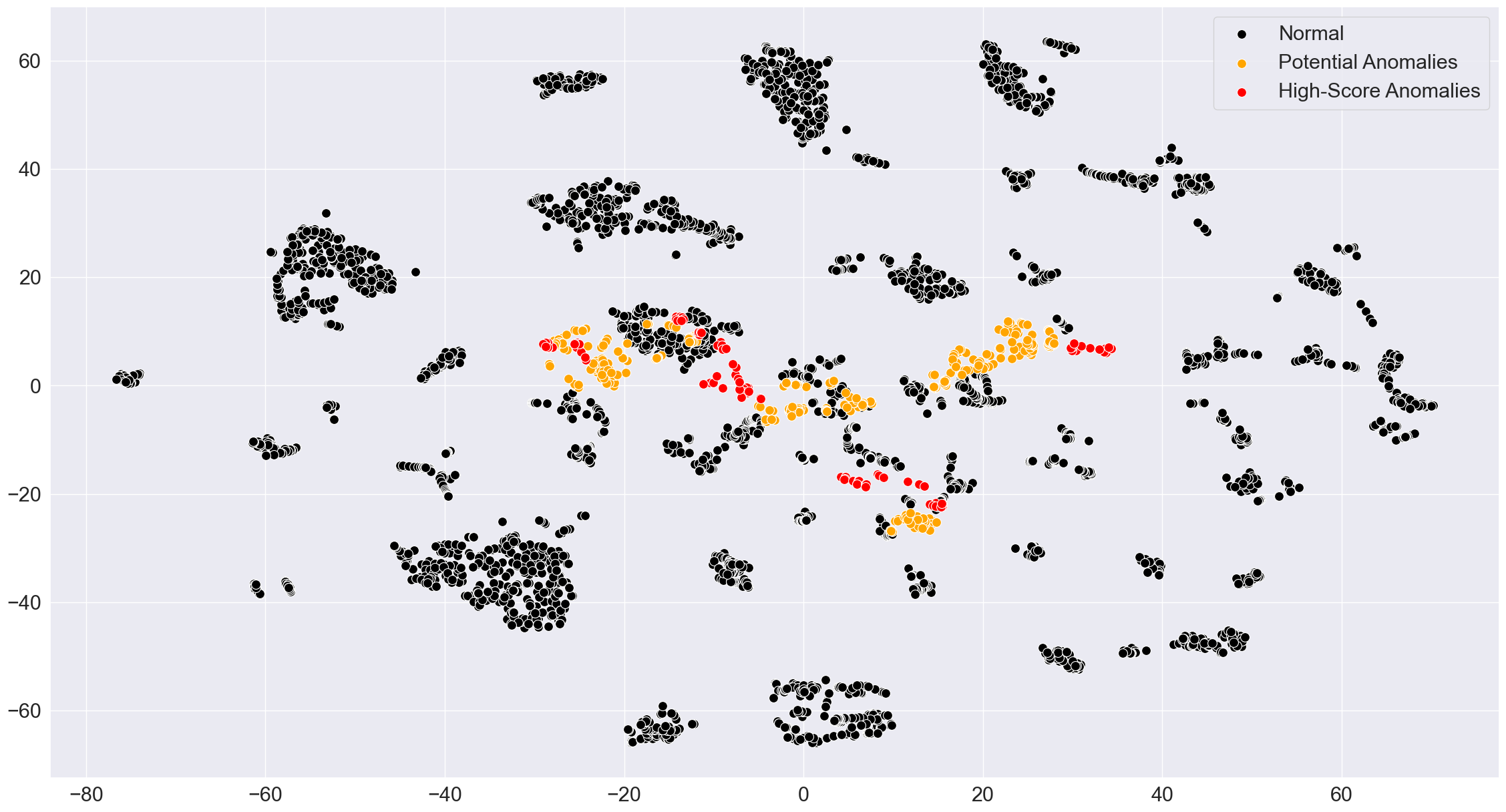}}}
\subfloat[Vanilla AE]{
    \label{fig:Test2_Vanilla_tSNE}
    \framebox{\includegraphics[width=0.48\linewidth, height=5cm]{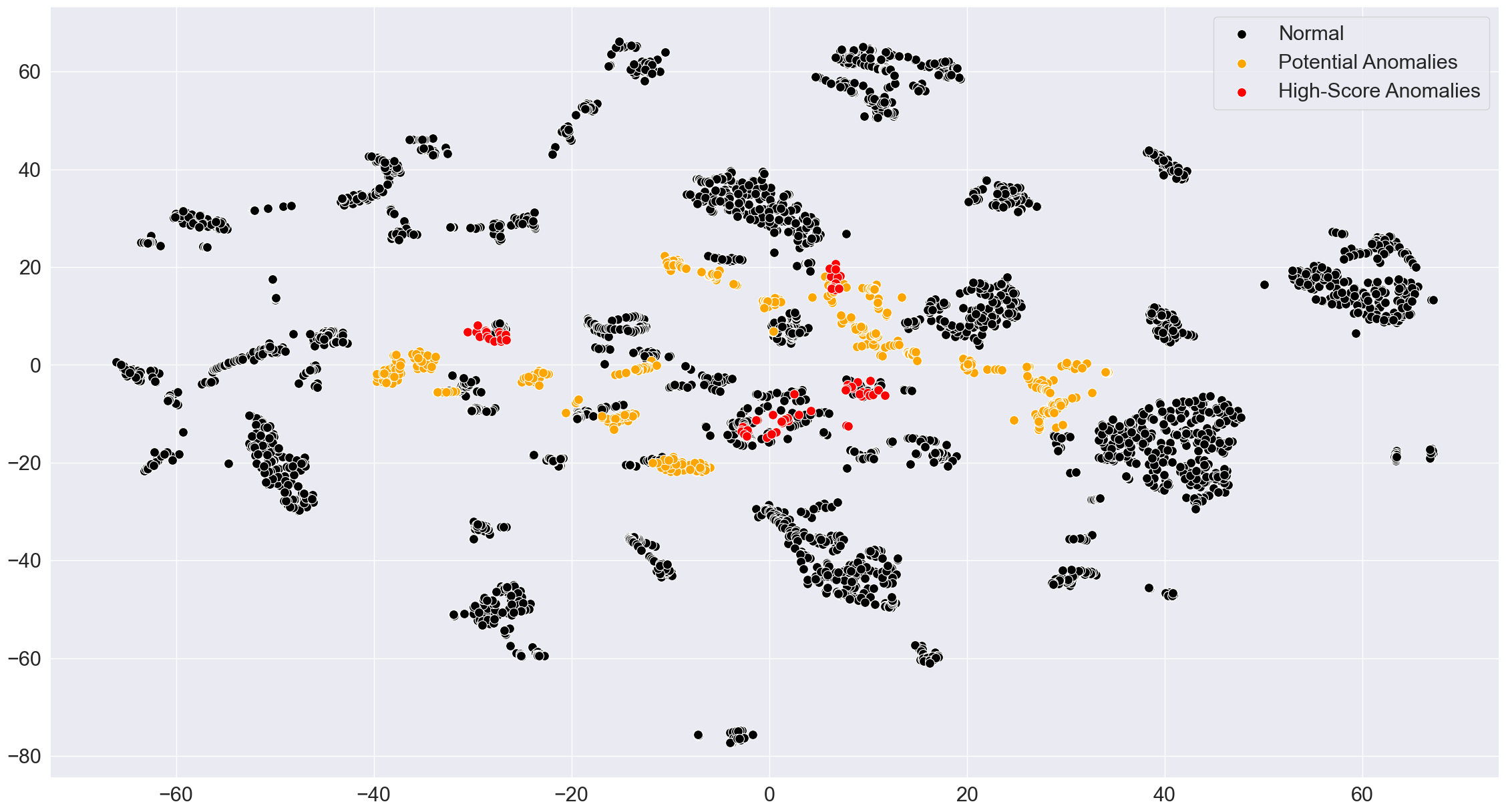}}}
\caption{t-SNE map of the test data 2 illustrating the inference result of the DL models.}
\label{fig:Test2_tSNE}
\end{figure*}

Figure \ref{fig:Test2_DL_anomaly_detection_models} shows the operational anomalies detected by the DL anomaly detection models on test data 2. Thus, Figure \ref{fig:Test2_tSNE} presents the t-SNE visualization of the test data 2 reflecting the inference results produced by the DL models.


\textit{Performance Evaluation.} The purpose of anomaly detection is to detect rare and critical events, so accuracy is not a metric emphasized in semi-supervised anomaly detection, since anomalies are rare compared to normal observations and the model is trained on normal data.
Then in this industrial experience study, the performance is discussed in terms of \textit{precision} and \textit{recall}, which indicate how many of the detected anomalies are actually correct (precision), and also how many of the known anomalous situations are detected properly (recall). The results from the inference of DL models on the test data show that the LSTM AE model performs a more precise detection. 
In particular, a detection precision $85\%$ and recall $91\%$ are reported for the LSTM AE anomaly detection model, while the vanilla AE shows notably lower performance. Moreover, the lower performance of the vanilla AE model can also be further shown by t-SNE maps in some instances, e.g., in test data 1 (Figure \ref{fig:Test1_Vanilla_tSNE}) the occurrence of the anomalous situations takes place at the beginning of the period and lasts only for a few days, so the detection of high-score anomalies over the whole span, especially at the end of the period, by the vanilla AE is not acknowledged.

\subsection{RQ2: Surrogate Interpretable Models}
\begin{figure*}[h]
\centering
\subfloat[Test data 1]{
    \label{fig:Test1_LSTM_Annotated_DT}
    \framebox{\includegraphics[height=1.8\textwidth, width = 5.5cm]{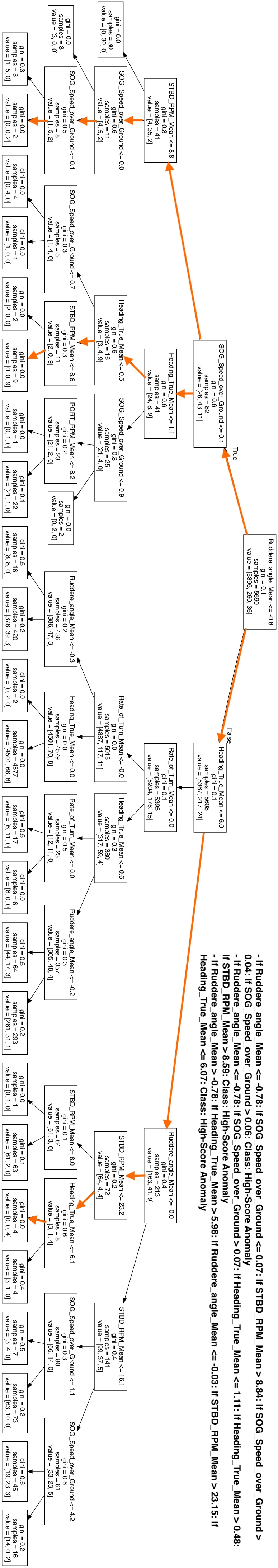}} } 
\subfloat[Test data 2]{
    \label{fig:Test2_LSTM_Annotated_DT}
    \framebox{\includegraphics[height =1.8\textwidth, width = 5.5cm]{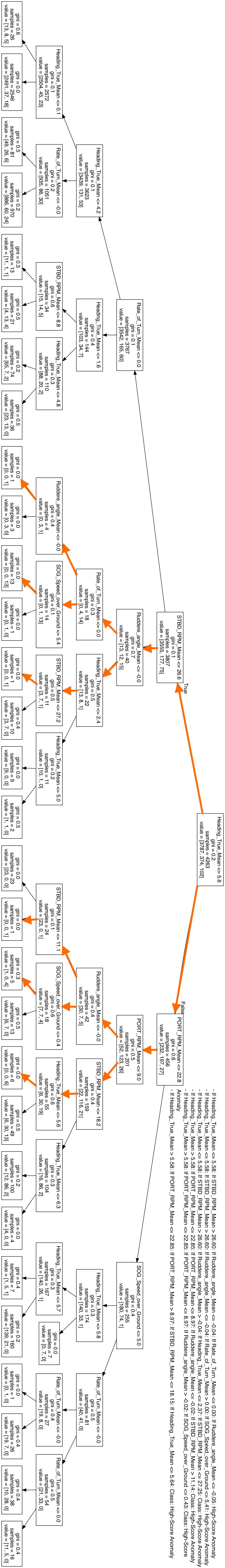}}}
\caption{Interpretable surrogate decision trees for the output of LSTM AE}
\label{fig:side_by_side_decision_trees}
\end{figure*}

Figures \ref{fig:Test1_LSTM_Annotated_DT} and \ref{fig:Test2_LSTM_Annotated_DT} present the surrogate DTs generated from the inference result of the LSTM AE on the test data 1 and 2. The depth of the DT model is adjustable so that the engineer can get a detailed view of the inference logic. Accordingly, a set of rules expressing the logic can also be extracted for the DT model. The rules are shown by the paths traversing from the root towards the leaf nodes in the DTs that present a decent homogeneity of the class distribution, $Gini\ Index = 0$. The Gini Index measures the impurity or diversity of the node's class distribution. A low Gini Index indicates a more homogeneous distribution of classes in the node and a better rule-indicating node. Tables \ref{table:Test1_Rules} and \ref{table:Test2_Rules} present a set of rules extracted from the pruned DT models on the test data 1 and 2. As per the engineer's review of the rules, it is noted that the rules presenting the inference by the LSTM AE model prove to be coherent and align with the expert's thought.

\begin{table*}
\centering
\caption{Test Data 1: Extracted rules from the generated decision trees}
\begin{tabular}{ |m{0.25\linewidth}|m{0.7\linewidth}|}
\hline
\textbf{DL Anomaly Detection Models } & \textbf{RULES} \\
\hline
\multirow{3}{*}\textbf{LSTM AE} & \vspace{2mm} 1) If Ruddere\_angle\_Mean $\leq -0.78$: If SOG\_Speed\_over\_Ground $\leq 0.07$: If STBD\_RPM\_Mean $> 8.84$: If SOG\_Speed\_over\_Ground $> 0.04$: If SOG\_Speed\_over\_Ground $> 0.06$: Class: High\textendash Score Anomaly \vspace{2mm} \\\cline{2-2}

&  \vspace{2mm} 2) If Ruddere\_angle\_Mean $\leq -0.78$: If SOG\_Speed\_over\_Ground $> 0.07$: If Heading\_True\_Mean $\leq  1.11$: If Heading\_True\_Mean $> 0.48$: If STBD\_RPM\_Mean $> 8.59$: Class: High\textendash Score Anomaly \vspace{2mm} \\\cline{2-2}

\textbf{} & \vspace{2mm} 3) If Ruddere\_angle\_Mean $> -0.78$: If Heading\_True\_Mean $> 5.98$: If Ruddere\_angle\_Mean $\leq -0.03$: If STBD\_RPM\_Mean $> 23.15$: If Heading\_True\_Mean $ \leq 6.07$: Class: High\textendash Score Anomaly \vspace{2mm}\\
\hline
\hline
\end{tabular}
\label{table:Test1_Rules}
\end{table*}

\section{Related Work}\label{Sec:Related Work}
In the context of maritime applications, where safety holds paramount importance, the integration of innovations in connection to the usage of data-driven techniques into existing vessels is a deliberative process. This is intricately linked to stringent regulations overseen by international bodies like the International Maritime Organization (IMO). Concurrently, the rise of autonomous systems in maritime operations necessitates a balanced approach that harmonizes safety concerns with technological progress. 



Monitoring vessel subsystems and gathering substantial volumes of data to facilitate continuous observation of vessel conditions and enable intelligent decision-making is a growing trend in the maritime industry, driven by advancements in IoT, data analytics, and automation technologies. Leveraging the NMEA2000 network to acquire time-series data from diverse sensors interconnected with on-board components has been extensively expounded upon in existing literature \cite{srecko2013, beirami2015, s19204480, Kessler2021}. The augmentation of this foundational framework through the integration of a software layer empowered by data analysis algorithms, i.e., different machine learning techniques can bring new functionalities, improving operations, crew decision-making, and safety. 
As it has offered pathways for leveraging data analysis approaches using various machine learning techniques to efficiently and effectively detect functional and performance anomalies \cite{6836578, 9190533, 9237642, 9917553, li2023novel, han2021fault}. In \cite{6836578}, a classical supervised ML method, a support vector machine, is trained on the labeled historic motion data and used to detect abnormal vessel motions. Singh and Heyman \cite{9190533} present how an SVM and a fully connected neural network can help with detecting AIS (Automatic Identification System) on-off switching anomalies. Li et al. \cite{9917553} propose an ensemble of DBSCAN clustering and an LSTM model for anomaly detection based on AIS data for vessel trajectories. Han et al. \cite{han2021fault} benefit from LSTM variational AE for fault detection (abnormal functional status detection) of a maritime diesel engine operating in a research-purpose vessel. In \cite{brandsaeter2016application} Auto Associative Kernel Regression (AAKR)---a nonparametric multivariate method used for signal reconstruction---along with a  Sequential Probability Ratio Test (SPRT), is employed for anomaly detection based on sensory data. SPRT is a statistical test used for sequential analysis in scenarios where data is collected sequentially, and decisions need to be made at each step. In this study, a few simple examples of synthetically generated anomalies are used to evaluate the performance of the proposed approach.   

Furthermore, the application of machine learning-enriched anomaly detection via multi-parameter observation also assumes a pivotal role in addressing the detection of cyber security threats \cite{8726823, 9141222} by assessing vessel conditions and identifying potential threats. It can bolster the resilience of maritime operations in the face of evolving cyber threats, contributing to a secure operational environment. 

\begin{table*}
\centering
\caption{Test Data 2: Extracted rules from the generated decision trees}
\begin{tabular}{ |m{0.25\linewidth}|m{0.7\linewidth}|}
\hline
\textbf{DL Anomaly Detection Models } & \textbf{RULES} \\
\hline
\multirow{6}{*}\textbf{LSTM AE} & \vspace{2mm} 1)
If Heading\_True\_Mean $\leq 5.58$: 
If STBD\_RPM\_Mean $> 26.60$: 
If Ruddere\_angle\_Mean $\leq -0.04$: 
If Rate\_of\_Turn\_Mean $\leq 0.00$: 
If Ruddere\_angle\_Mean $\leq -0.05$: \text{Class: High-Score Anomaly} \vspace{2mm} \\\cline{2-2} 

& \vspace{2mm} 2) If Heading\_True\_Mean $\leq 5.58$: 
If STBD\_RPM\_Mean $> 26.60$: \text{If Ruddere\_angle\_Mean} $\leq -0.04$: 
\text{If Rate\_of\_Turn\_Mean} $> 0.00$: 
\text{If SOG\_Speed\_over\_Ground} $\leq 5.41$: \text{Class: High-Score Anomaly} 
\vspace{2mm} \\\cline{2-2} 

& \vspace{2mm} 3) If Heading\_True\_Mean $\leq 5.58$: 
If STBD\_RPM\_Mean $> 26.60$: \text{If Ruddere\_angle\_Mean} $> -0.04$: 
\text{If Heading\_True\_Mean} $\leq 2.37$: 
\text{If STBD\_RPM\_Mean} $\leq 27.25$: \text{ Class: High-Score Anomaly} 
\vspace{2mm} \\\cline{2-2} 

& \vspace{2mm} 4) \text{If Heading\_True\_Mean} $> 5.58$: 
\text{If PORT\_RPM\_Mean} $\leq 22.85$: 
\text{If PORT\_RPM\_Mean} $\leq 8.97$: 
\text{If Ruddere\_angle\_Mean} $\leq -0.02$: 
\text{If STBD\_RPM\_Mean} $> 11.14$: \text{ Class: High-Score Anomaly} 
\vspace{2mm} \\\cline{2-2}

& \vspace{2mm} 5) \text{If Heading\_True\_Mean} $> 5.58$: \text{If PORT\_RPM\_Mean} $\leq 22.85$: 
\text{If PORT\_RPM\_Mean} $\leq 8.97$: 
\text{If Ruddere\_angle\_Mean} $> -0.02$: 
\text{If SOG\_Speed\_over\_Ground} $\leq 0.43$: \text{ Class: High-Score Anomaly} 
\vspace{2mm} \\\cline{2-2}

& \vspace{2mm} 6) \text{If Heading\_True\_Mean} $> 5.58$: \text{If PORT\_RPM\_Mean} $\leq 22.85$: 
\text{If PORT\_RPM\_Mean} $> 8.97$: 
\text{If STBD\_RPM\_Mean} $\leq 18.15$: 
\text{If Heading\_True\_Mean} $\leq 5.64$: \text{ Class: High-Score Anomaly} \vspace{2mm} \\
\hline
\hline
\end{tabular}
\label{table:Test2_Rules}
\end{table*}

\section{Conclusion}\label{Sec:Conclusion}
{This industry-applied study presents the application of an ML-driven operational anomaly detection framework on the operational data of the sensorized vessel TUCANA. It showcases the practical integration of deep learning models with interpretable solutions to meet industry needs for transparency and efficiency in the investigation of anomalous situations. The use of the LSTM AE model enables the effective identification of operational anomalies and the integration of interpretable surrogate models provides the engineers with meaningful insights into the reasoning behind the detections. They not only prove practicable but also align closely with the engineer's perspectives, enhancing the interpretability and trustworthiness of the anomaly detection system. The incorporation of t-SNE as a baseline unsupervised anomaly detection method provides an additional layer of visualization for engineers, allowing them to better assess the distribution of identified anomalies relative to underlying clusters. The empirical evaluation of the framework on real-world test datasets, including scenarios involving propeller failures and critical maneuvers, highlights its effectiveness and practical applicability for industrial anomaly detection in a maritime use case.

Investigating adaptive thresholding methods to better handle evolving operational conditions and reduce false positives and utilizing hybrid Model architecture remain a direction for future works. We plan to also study the potential of federated learning across a number of functional vessels to enable an anomaly detection model generalizing across different vessels or operational scenarios while preserving data privacy.

\section*{Acknowledgment}
{This work has been funded by H2020 ECSEL Joint Undertaking (JU) InSecTT (\url{https://www.insectt.eu/}) and TRANSACT (\url{https://transact-ecsel.eu//}) projects under grant agreement No 876038 and No 101007260 respectively.}

\bibliographystyle{IEEEtran}
\bibliography{TUCANA_}

\end{document}